\documentclass[final]{article}

\usepackage{wrapfig}
\usepackage{hyperref}
\usepackage[american]{babel}
\usepackage{microtype}

\usepackage[nonatbib]{nips_2018}

\usepackage[utf8]{inputenc} 
\usepackage[T1]{fontenc}    
\usepackage{hyperref}       
\usepackage{url}            
\usepackage{booktabs}       
\usepackage{graphicx}
\usepackage{amsfonts}       
\usepackage{nicefrac}       
\usepackage{microtype}      
\usepackage{CJKutf8}
\AtBeginDvi{\input{zhwinfonts}}
\usepackage{amsmath}
\usepackage[linesnumbered,ruled,vlined]{algorithm2e}
\usepackage{multicol}
\usepackage{multirow}

\title{Product Title Refinement via Multi-Modal Generative Adversarial Learning}

\usepackage{authblk} 
\author{
  \textbf{Jian-Guo Zhang,}$^1$ \textbf{Pengcheng Zou,}$^2$ \textbf{Zhao Li,}$^2$  \textbf{Yao Wan,}$^3$ \textbf{Ye Liu,}$^1$  \hspace{3cm} \textbf{Xiuming Pan,}$^2$  \textbf{Yu Gong,}$^2$ \textbf{Philip S. Yu}$^1$ \\
  $^1$Department of Computer Science, University of Illinois at Chicago, Illinois, USA\\
  $^2$Alibaba Group\\
  $^3$College of Computer Science and Technology, Zhejiang University, Hangzhou, China\\
  \texttt{\{jzhan51,yliu279,psyu\}@uic.edu}, \texttt{wanyao@zju.edu.cn},  \texttt{\{xuanwei.zpc,lizhao.lz,xuming.panxm,gongyu.gy\}@alibaba-inc.com} 
}

\begin{document}

\maketitle

\begin{abstract}
Nowadays, an increasing number of customers are in favor of using E-commerce Apps to browse and purchase products. Since merchants are usually inclined to employ redundant and over-informative product titles to attract customers' attention, it is of great importance to concisely display short product titles on limited screen of cell phones. Previous researchers mainly consider textual information of long product titles and lack of human-like view during training and evaluation procedure. In this paper, we propose a Multi-Modal Generative Adversarial Network (MM-GAN) for short product title generation, which innovatively incorporates image information, attribute tags from the product and the textual information from original long titles. MM-GAN treats short titles generation as a reinforcement learning process, where the generated titles are evaluated by the discriminator in a human-like view.

\end{abstract}

\section{Introduction}
E-commerce companies such as TaoBao and Amazon put many efforts to improve the user experience of their mobile Apps. 
For the sake of improving retrieval results by search engines, merchants usually write lengthy, over-informative, and sometimes incorrect titles, e.g., the original product title may contain more than 20 Chinese words, which may be suitable for PCs. However, these titles are cut down and no more than 10 words can be displayed on a mobile phone with limited screen size varying from 4 to 5.8 inches. Hence, to properly display products in mobile screen, it is important to produce succinct short titles to preserve important information of original long titles and accurate descriptions of products.


This problem is related to text summarization, which can be categorized into two classes: extractive \cite{cao2016attsum,nallapati2017summarunner,miao2016language}, and abstractive \cite{chopra2016abstractive,chen2016distraction,see2017get} methods. The extractive methods select important words from original titles, while the abstractive methods generate titles by extracting words from original titles or generating new words from data corpus. They usually approximate such goals by predicting the next words given previous predicted words using maximum likelihood estimation (MLE) objective. Despite their success to a large extent, they suffer from the issue of exposure bias \cite{ranzato2015sequence}. It may cause the models to behave in undesired ways, e.g., generating repetitive or truncated outputs. In addition, predicting next words based on previously generated words will make the learned model lack of human-like holistic view of the whole generated short product titles.

More recent state-of-the-art methods \cite{gong2018automatic,wang2018multi} treat short product titles generation as a sentence compression task following attention-based extractive mechanism. They extract key characteristics mainly from original long product titles. 
However, in real E-Commerce scenario, product titles are usually redundant and over-informative, and sometimes even inaccurate.  e.g., long titles of a cloth may include both
\begin{CJK}{UTF8}{gbsn}
``嘻哈$|$狂野 (hip-pop$|$wild)" and ``文艺$|$淑女 (artsy$|$delicate)"
\end{CJK}
simultaneously. It is very hard to generate succinct and accurate short titles just relying on the original titles. Therefore, it is insufficient to regard short title generation as traditional text summarization problem in which original text has already contained complete information.


In this paper,
we propose a novel \textbf{M}ulti-\textbf{M}odal \textbf{G}enerative \textbf{A}dversarial \textbf{N}etwork, named \textbf{MM-GAN}, to better generate short product titles. It includes a generator and a discriminator. The generator generates a short product title based on original long titles, while additional information from corresponding visual image and attribute tags, the discriminator distinguishes whether the generated short titles are human-produced or machine-produced in a human-like view. The task is treated as a reinforcement learning problem, in which the quality of a machine-generated short product title depends on its ability to fool the discriminator into believing it is generated by human, and output of the discriminator is a reward for the generator to improve generated quality.

We highlight that our model can: (a) incorporate the image and attribute tags aside from original long product titles into the generator. To the best of our knowledge, it's the first attempt for us to design a multi-modal model to consider multiple modalities of inputs for better short product titles generation in E-commerce;  
(b) generate short product titles in a holistic manner.


\section{Multi-Modal Generative Adversarial Network}


In this section, we describe in details the proposed MM-GAN. The problem can be formulated as follows: given an original long product title $L=\left \{l_1, l_2, ..., l_K \right\}$
consisted of $K$ Chinese or English words, a single word can be represented in a form like ``skirt" in English or \begin{CJK}{UTF8}{gbsn}``半身裙"\end{CJK} in Chinese. With an additional image $I$ and attribute tags $A=\left \{a_1, a_2, ..., a_M \right\}$, the model targets at generating a human-like short product title $S= \left \{s_1, s_2, ...,s_N\right \}$, where $M$ and $N$ are the number of words in $A$ and $S$, respectively.  


\subsection{Multi-Modal Generator}
\begin{wrapfigure}{r}{0.5\textwidth}
    \centering
    \includegraphics[width=0.5\textwidth]{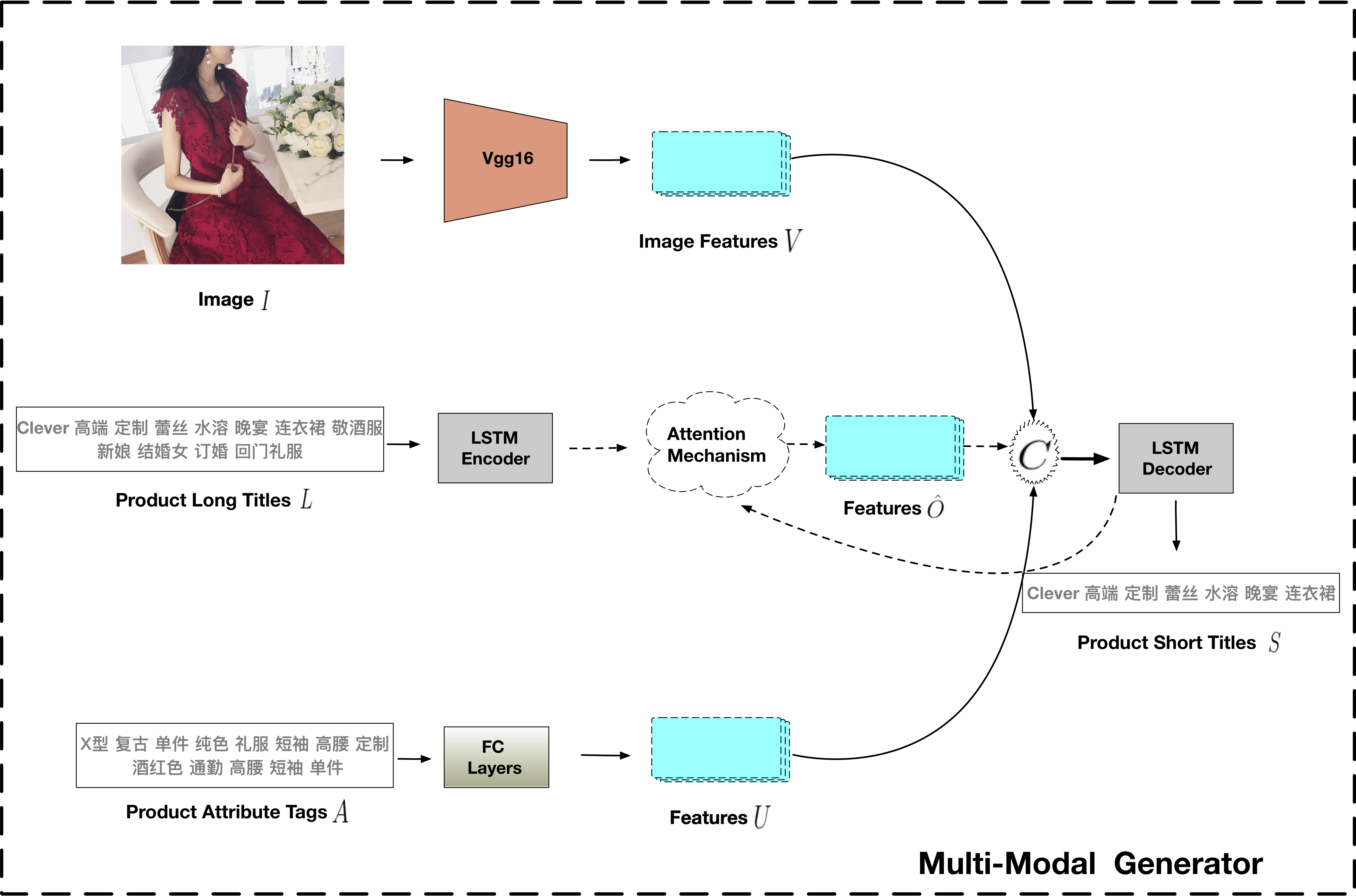}
    \caption{Multi-Modal Generator of MM-GAN.}
    \label{fig:framework1}
\end{wrapfigure}

The multi-modal generative model defines a policy of generating short product titles $S$ given original long titles $L$, with additional information from product image $I$ and attribute tags $A$. Fig. \ref{fig:framework1} illustrates the architecture of our proposed multi-modal generator which follows the seq2seq \cite{sutskever2014sequence} framework.
\textbf{Multi-Modal Encoder. }
As we mentioned before, our model tries to incorporate multiple modalities of a product (i.e., image, attribute tags and long title). To learn the multi-modal embedding of a product, we first adopt a pre-trained VGG16 \cite{simonyan2014very} as the CNN architecture to extract features $V=[v_1, v_2, ..., v_Z]$ of an image $I$ from the condensed fully connected layers, where $Z$ is the number of latent features. In order to get more descriptive features, we fine-tune the last 3 layers of VGG16 based on a supervised classification task given classes of products images. 
Second, we encode the attribute tags $A$ into a fixed-length feature vector $U=\left [u_1, u_2, \ldots, u_M\right ]$, and $U=f_1(A)$, where $f_1$ denotes fully connected layers.
Third, we encode the original long titles $L$. Specifically, the features extracted from original long titles $L$ are $O=\left [o_1, o_2, \ldots, o_K \right]$, where $o_t=f_2(o_{t-1}, l_t)$. Here $f_2$ represents a non-linear function, and in this paper the LSTM unit \cite{hochreiter1997long} is adopted.

\textbf{Decoder.}
The hidden state $h_t$ for the $t$-th target word $s_t$ in short product titles $S$ can be calculated as $h_t=f_2(h_{t-1},s_{t-1}, \hat{o}_t)$.
Here we adopt an attention mechanism \cite{bahdanau2014neural} to capture important words from original long titles $L$.
The context vector $\hat{o}_t$ is a weighted sum of hidden states $O$, it can be computed by
$\hat{o}_t=\sum_{k=1}^{K}\alpha_{t,k}o_k$,
where $\alpha_{k,t}$ is the contribution of an input word $l_k$ to the $t$-th target word using an alignment model $a\_m$ \cite{bahdanau2014neural}:
$\alpha_{t,k}=\frac{\exp(a\_m(h_{t-1}, o_k))}{\sum_{k'=1}^{K}\exp(a\_m(h_{t-1}, o_k'))}$.
After obtaining all features $U$, $V$, $\hat{O}$ from $A$, $I$ and $L$, respectively, we then concatenate them into the final feature vector:
$C=\tanh(W[\hat{O};V;U])$,
where $W$ are learnable weights and $[;]$ denotes the concatenation operator. 
Finally, $C$ is fed into the LSTM based decoder to predict the probability of generating each target word for short product titles $S$. As the sequence generation problem can be viewed as a sequence decision making process \cite{bachman2015data}, we denote the whole generation process as a policy $\pi(S|C)$.

\subsection{Discriminator}
The discriminator model $D$ is a binary classifier which takes an input of a generated short product titles $S$ and distinguishes whether it is human-generated or machine-generated. The short product titles are encoded into a vector representation through a two-layer LSTM model, and sent to a two-way softmax function, which returns the probability of the input short product titles being generated by human:
$P=softmax(W_d\left [ LSTM(S)\right ]+b_d)$, where $W_d$ is a weight matrix and $b_d$ is a bias.  

\subsection{End-to-End Training}

The multi-modal generator $G$ tries to generate a sequence of tokens $S$ under a policy $\pi$ and fool the discriminator $D$ via maximizing the reward signal received from $D$. The objective of $G$ can be formulated as follows:
\begin{equation}
    J(\theta)=\mathbb{E}_{S\sim \pi_{\theta}\left ( S|C \right )}\left [ R_{\phi }(S) \right ],
\end{equation}
where $\theta$ and $\phi$ are learnable parameters for $G$ and $D$, respectively. 

Conventionally, GANs are designed for generating continuous data and thus $G$ is differential with continuous parameters guided by the objective function from $D$ \cite{yu2017seqgan}. Unfortunately, it has difficulty in updating parameters through back-propagation when dealing with discrete data in text generation. To solve the problem, we adopt the REINFORCE algorithm \cite{williams1992simple}. Specifically, once the generator reaches the end of a sequence (i.e., $S=S_{1:T}$), it receives a reward $R_{\phi}\left ( S \right )$ from $D$ based on the probability of being real. 


In text generation, $D$ will provide a reward to $G$ only when the whole sequence has been generated, and no intermediate reward is obtained before the final token of $S$ is generated. This may cause the discriminator to assign a low reward to all tokens in the sequence though some tokens are proper results. To mitigate the issue, we utilize Monte Carlo (MC) search with $N'$-time roll-outs \cite{yu2017seqgan} to assign rewards to intermediate tokens.  
$\left \{ S_{1:N}^1, \ldots, S_{1:N}^{N'} \right \}=MC^{\pi}\left ( S_{1:N}; N'\right )$
where $S_{1:t}^n=\left ( s_1, \ldots, s_t \right )$ and $S_{t+1:T}^n$ are sampled based on roll-out policy $\pi$ and the current state. The intermediate reward now is $R_{\phi}(S_{1:t-1}, a'=s_t)=\frac{1}{N'}\sum_{n=1}^{N'}R(S_{1:N}^n)$,
here $a'$ is an action at current state $t$. Now we can compute the gradient of the objective function for the generator $G$:
\begin{align} \label{eq-generator}
      & \nabla_{\theta} J(\theta) 
      \approx\mathbb{E}_{S\sim \pi_{\theta}\left ( S|C \right )}\left [ \sum_{t=1}^{N} \pi_{\theta \left ( s_t|C \right )}\nabla_{\theta}\log(\pi_{\theta\left ( s_t|C \right )})R_{\phi}(s_{1:t})\right ],
\end{align}
where $\nabla_{\theta}$ is the partial differential operator for $\theta$ in $G$, and $R_{\phi}$ is fixed during updating of generator. 

The objective function for the discriminator $D$ can be formulated as:
\begin{equation}\label{object-function}
    \mathbb{E}_{S\sim \mathcal{P}_{\theta}\left ( S|C \right )}\left [ \log R_{\phi}(S|C) \right ]-\mathbb{E}_{S\sim \pi_{\theta}\left ( S|C \right )}\left [ \log R_{\phi}(S|C) \right ].
\end{equation}

\section{Experiments}

\subsection{Experimental Setup}

\textbf{Dataset. }We train our model on LESD4EC dataset \cite{gong2018automatic}, which consists of more than 6M products from Taobao Platform, each product includes a long product title and a short product title written by professional writers, along with a high quality image and attributes tags. We randomly select $80\%$ of data as training set, and the rest  as validation set and test set ($10\%$ each).

\textbf{Baselines.} We compare our proposed model with the following four baselines: (a) Pointer Network (\textbf{Ptr-Net}) \cite{see2017get} which is a seq2seq based framework with pointer-generator network copying words from the source text via \textit{pointing}.
(b) Feature-Enriched-Net (\textbf{FE-Net}) \cite{gong2018automatic} which is a deep and wide model based on attentive RNN to generate the textual long product titles. 
(c) Agreement-based MTL (\textbf{Agree-MTL}) \cite{wang2018multi} which is a multi-task learning approach to improve product title compression with user searching log data.
(d) Generative Adversarial Network (\textbf{GAN}) \cite{li2017adversarial} which is a generative adversarial method for text generation with only single input. 

\subsection{Automatic Evaluation}
To evaluate the quality of generated product short titles, we follow \cite{wang2018multi,gong2018automatic} and use standard recall-oriented ROUGE metric \cite{lin2004rouge}. 
Experimental results on test set are shown in Table \ref{tab-table2}. From this table, we note that our proposed MM-GAN achieves best performance on three metrics. Furthermore, when comparing MM-GAN with GAN, we can find that additional information such as image and attribute tags from product can absolutely facilitate our model to generate better short titles. 
In addition, our proposed model also outperforms Agree-MTL which can be illustrated from two aspects: 
(a) it incorporates multiple sources, containing more information than other single-source based model.
(b) it applies a discriminator to distinguish whether a product short titles are human-generated or machine-generated, which makes the model evaluate the generated sequence in a human-like view, and naturally avoid exposure bias in other methods. 



\begin{table}
\centering
\begin{tabular}{l|cccccc}
Models     & Ptr-Net & FE-Net & Agree-MTL & GAN & MM-GAN  \\ \hline
ROUGE-1  & 59.21   & 61.07  & 66.19     & 60.67       &\textbf{69.53}    \\
ROUGE-2   & 42.01   & 44.16  & 49.39     & 46.46       &\textbf{52.38}  \\
ROUGE-L & 57.12   & 58.00  & 64.04     & 60.27       &\textbf{65.80} 
\end{tabular}
\caption{ROUGE performance of different models on the test set.}
\label{tab-table2}
\end{table}

\begin{table}[]
\resizebox{\textwidth}{!}{%
\begin{tabular}{|cc|l|l|}
\hline
\multicolumn{2}{|c|}{Data}                                          & \multicolumn{1}{c|}{Methods} & \multicolumn{1}{c|}{Results} \\ \hline
Product Long Titles                         & \multicolumn{1}{l|}{\begin{tabular}[c]{@{}l@{}} \begin{CJK}{UTF8}{gbsn}Artka 阿卡 夏新 花边 镂空 荷叶边 抽绳 民族 狂野 复古 衬衫 S110061Q\end{CJK}\\ (Artka Artka summer lace hollow-out flounce drawstring \\ nation wild retro shirt S110061Q) \end{tabular} } & FE-Net                       &      \begin{tabular}[c]{@{}l@{}} \begin{CJK}{UTF8}{gbsn}阿卡 花边 镂空 荷叶边  衬衫\end{CJK} \\ (Artka lace hollow-out flounce  shirt) \end{tabular}                            \\ \hline
\multicolumn{1}{|c|}{\multirow{3}{*}{Image \begin{minipage}{0.3\textwidth}
      \includegraphics[width=0.5\linewidth]{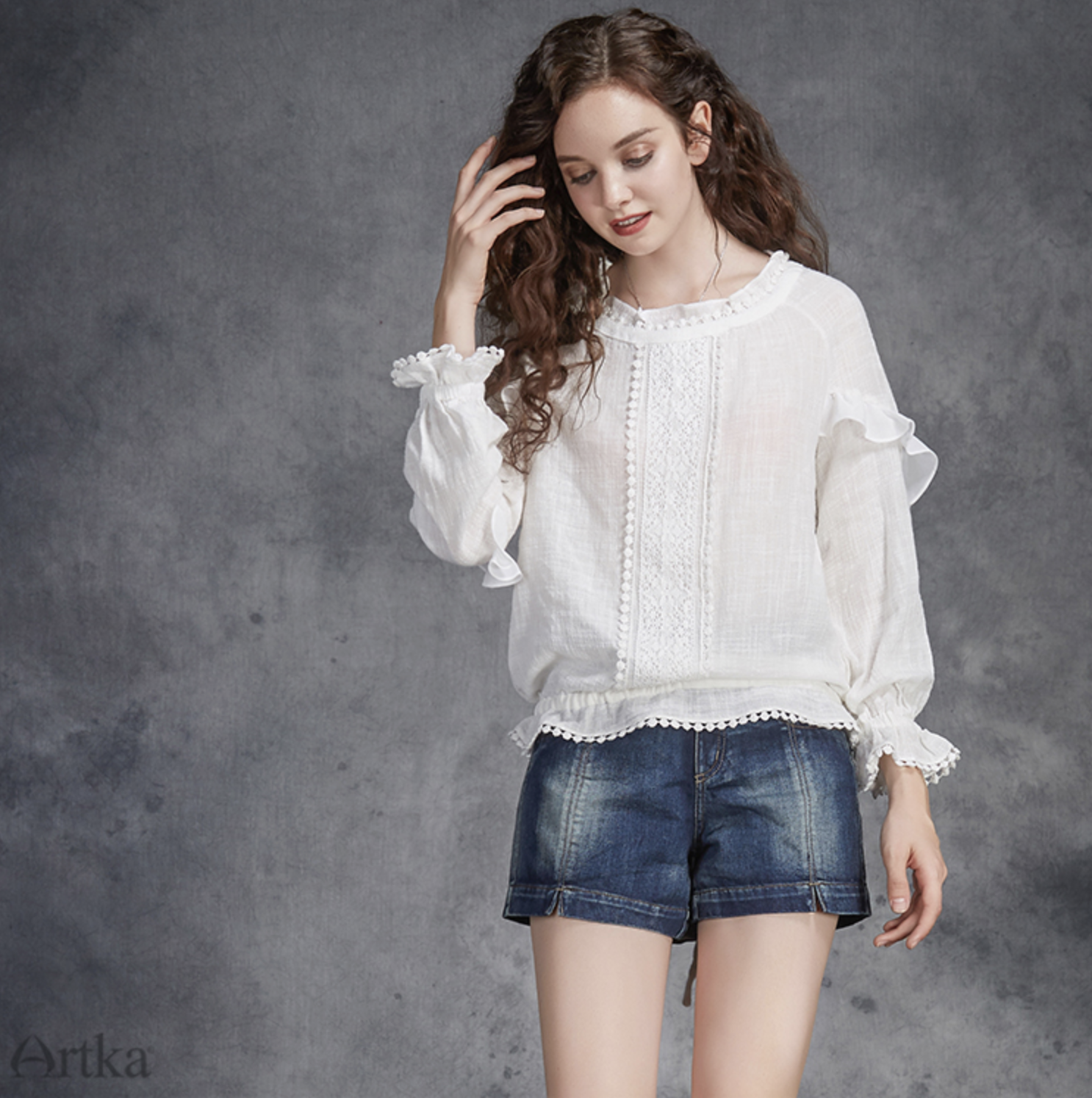}
    \end{minipage}}} & \multirow{3}{*}{Attributes Tags  \begin{tabular}[c]{@{}l@{}} \begin{CJK}{UTF8}{gbsn}修身 常规款 圆领  Artka 米白 长袖 \end{CJK} \\ \begin{CJK}{UTF8}{gbsn}套头 复古  通勤 纯色 夏季 喇叭袖 棉\end{CJK}  \\ (slim common round-neck Artka off-white long-sleeve \\ pullover retro commuting plain summer flare-sleeve cotton)\end{tabular} }     & Agree-MTL                    &             \begin{tabular}[c]{@{}l@{}} \begin{CJK}{UTF8}{gbsn}Artka 阿卡 夏新 花边 镂空 荷叶边 衬衫\end{CJK} \\ (Artka Artka summer lace hollow-out flounce  shirt) \end{tabular}                    \\ \cline{3-4} 
\multicolumn{1}{|c|}{}                      &                       & GAN                  &                          \multicolumn{1}{l|}{\begin{tabular}[c]{@{}l@{}} \begin{CJK}{UTF8}{gbsn}Artka 荷叶边 抽绳 衬衫\end{CJK} \\ (Artka  lace  flounce drawstring shirt) \end{tabular} }                \\ \cline{3-4} 
\multicolumn{1}{|c|}{}                      &                       & MM-GAN                       &               \multicolumn{1}{l|}{\begin{tabular}[c]{@{}l@{}} \begin{CJK}{UTF8}{gbsn}Artka 花边 荷叶边 镂空 复古 衬衫\end{CJK} \\ (Artka  lace flounce hollow-out retro shirt) \end{tabular}}                 \\ \hline
\end{tabular}
}
\caption{The comparison of generated short titles among different methods.} \label{tab-case}
\end{table}

\subsection{Case Study}


Table \ref{tab-case} shows a sample of product short titles generated by MM-GAN and baselines.
From this table, we can note that (a) product short titles generated by our model are more fluent, informative than baselines, and core product words (e.g., \begin{CJK}{UTF8}{gbsn}``Artka$|$ (阿卡)", ``复古$|$ (retro)", ``衬衫$|$ (skirt)"\end{CJK}) can be recognized. (b) There are over-informative words (e.g., \begin{CJK}{UTF8}{gbsn}``阿卡$|$ (Artka)", ``S110061Q"\end{CJK}) and irrelevant words (e.g., \begin{CJK}{UTF8}{gbsn}``狂野$|$ (wild)"\end{CJK}) in product long titles. Over-informative words may disturb model's generation process, irrelevant words may give incorrect information to the model. These situations could happen in real E-commerce environment. FE-Net misses the English brand name ``Artka" and gives its Chinese name \begin{CJK}{UTF8}{gbsn}`阿卡"\end{CJK} instead. Agree-MTL using user searching log data performs better than GAN. However, Agree-MTL still generates the over-informative word \begin{CJK}{UTF8}{gbsn}`阿卡"\end{CJK}. MM-GAN outperforms all baselines, information in additional attribute tags such as \begin{CJK}{UTF8}{gbsn}``复古$|$ (retro)", ``Artka"\end{CJK}), and other information from the product main image are together considered by the model and help the model select core words and filter out irrelevant words in generated product short titles. Which shows that MM-GAN using different types of inputs can help generate better product short titles.

\section{Conclusion}

In this paper, we propose a multi-modal generative adversarial network for short product title generation in E-commerce. Different from conventional methods which only consider textual information from long product titles, we design a multi-modal generative model to incorporate additional information from product image and attribute tags. 
Extensive experiments on a large real-world E-commerce dataset verify the effectiveness of our proposed model when comparing with several state-of-the-art baselines.

\newpage
\bibliographystyle{plain}
\bibliography{reference}

\begin{thebibliography}{10}

\bibitem{bachman2015data}
Philip Bachman and Doina Precup.
\newblock Data generation as sequential decision making.
\newblock In {\em NIPS}, pages 3249--3257, 2015.

\bibitem{bahdanau2014neural}
Dzmitry Bahdanau, Kyunghyun Cho, and Yoshua Bengio.
\newblock Neural machine translation by jointly learning to align and
  translate.
\newblock In {\em ICLR}, 2015.

\bibitem{cao2016attsum}
Ziqiang Cao, Wenjie Li, Sujian Li, Furu Wei, and Yanran Li.
\newblock Attsum: Joint learning of focusing and summarization with neural
  attention.
\newblock In {\em COLING}, 2016.

\bibitem{chen2016distraction}
Qian Chen, Xiaodan Zhu, Zhenhua Ling, Si~Wei, and Hui Jiang.
\newblock Distraction-based neural networks for document summarization.
\newblock In {\em IJCAI}, 2016.

\bibitem{chopra2016abstractive}
Sumit Chopra, Michael Auli, and Alexander~M Rush.
\newblock Abstractive sentence summarization with attentive recurrent neural
  networks.
\newblock In {\em NAACL-HLT}, pages 93--98, 2016.

\bibitem{gong2018automatic}
Yu~Gong, Xusheng Luo, Kenny~Q Zhu, Shichen Liu, and Wenwu Ou.
\newblock Automatic generation of chinese short product titles for mobile
  display.
\newblock In {\em IAAI}, 2018.

\bibitem{hochreiter1997long}
Sepp Hochreiter and J{\"u}rgen Schmidhuber.
\newblock Long short-term memory.
\newblock {\em Neural computation}, 9(8):1735--1780, 1997.

\bibitem{li2017adversarial}
Jiwei Li, Will Monroe, Tianlin Shi, S{\'e}bastien Jean, Alan Ritter, and Dan
  Jurafsky.
\newblock Adversarial learning for neural dialogue generation.
\newblock In {\em EMNLP}, 2017.

\bibitem{lin2004rouge}
Chin-Yew Lin.
\newblock Rouge: A package for automatic evaluation of summaries.
\newblock {\em Text Summarization Branches Out}, 2004.

\bibitem{miao2016language}
Yishu Miao and Phil Blunsom.
\newblock Language as a latent variable: Discrete generative models for
  sentence compression.
\newblock In {\em EMNLP}, 2016.

\bibitem{nallapati2017summarunner}
Ramesh Nallapati, Feifei Zhai, and Bowen Zhou.
\newblock Summarunner: A recurrent neural network based sequence model for
  extractive summarization of documents.
\newblock In {\em AAAI}, pages 3075--3081, 2017.

\bibitem{ranzato2015sequence}
Marc'Aurelio Ranzato, Sumit Chopra, Michael Auli, and Wojciech Zaremba.
\newblock Sequence level training with recurrent neural networks.
\newblock In {\em ICLR}, 2016.

\bibitem{see2017get}
Abigail See, Peter~J Liu, and Christopher~D Manning.
\newblock Get to the point: Summarization with pointer-generator networks.
\newblock In {\em ACL}, 2017.

\bibitem{simonyan2014very}
Karen Simonyan and Andrew Zisserman.
\newblock Very deep convolutional networks for large-scale image recognition.
\newblock In {\em ICLR}, 2015.

\bibitem{sutskever2014sequence}
Ilya Sutskever, Oriol Vinyals, and Quoc~V Le.
\newblock Sequence to sequence learning with neural networks.
\newblock In {\em NIPS}, pages 3104--3112, 2014.

\bibitem{wang2018multi}
Jingang Wang, Junfeng Tian, Long Qiu, Sheng Li, Jun Lang, Luo Si, and Man Lan.
\newblock A multi-task learning approach for improving product title
  compression with user search log data.
\newblock In {\em AAAI}, 2018.

\bibitem{williams1992simple}
Ronald~J Williams.
\newblock Simple statistical gradient-following algorithms for connectionist
  reinforcement learning.
\newblock {\em Machine learning}, 8(3-4):229--256, 1992.

\bibitem{yu2017seqgan}
Lantao Yu, Weinan Zhang, Jun Wang, and Yong Yu.
\newblock Seqgan: Sequence generative adversarial nets with policy gradient.
\newblock In {\em AAAI}, pages 2852--2858, 2017.

\end{thebibliography}

\end{document}